\documentclass[aps,pra,superscriptaddress,twocolumn,longbibliography]{revtex4-1}
\usepackage{amsmath}
\usepackage{comment}
\usepackage[margin=1in]{geometry}
\usepackage{graphicx}
\usepackage{listings}
\usepackage{multirow}
\usepackage{subcaption}
\newcommand{\etal}{{\it et.\,al.~}}
\newcommand{\ie}{{\it i.\,e.~}}
\usepackage{xcolor}
\definecolor{maroon}{RGB}{100,20,20}
\definecolor{dblue}{RGB}{20,20,100}
\definecolor{deepgreen}{rgb}{0.0, 0.4, 0.0}
\usepackage[colorlinks=true,linkcolor=dblue,citecolor=maroon,urlcolor=maroon]{hyperref}
\usepackage{ulem}

\begin{document}
\title{Inferring physical laws by artificial intelligence based
causal models}
\author{Jorawar Singh}
\email{ph19023@iisermohali.ac.in}
\affiliation{Department of Physical Sciences, Indian
Institute of Science Education and Research (IISER) Mohali,
Sector 81 SAS Nagar, Manauli PO 140306 Punjab, India}
\author{Kishor Bharti}
\email{kishor.bharti1@gmail.com}
\affiliation{Institute of High Performance Computing (IHPC),
Agency for Science, Technology and Research (A*STAR),
1 Fusionopolis Way, $\#$16-16 Connexis, Singapore 138632,
Republic of Singapore}
\author{Arvind}
\email{arvind@iisermohali.ac.in }
\affiliation{Department of Physical Sciences, Indian
Institute of Science Education and Research (IISER) Mohali,
Sector 81 SAS Nagar, Manauli PO 140306 Punjab, India}
\affiliation{Punjabi University, Patiala,
147002, Punjab, India}
\begin{abstract}
The advances in  Artificial Intelligence (AI) and Machine
Learning (ML) have  opened up many avenues for scientific 
research, and are adding new dimensions to the process of 
knowledge creation. However, even the most powerful and 
versatile of ML applications till date are primarily in the
domain of analysis of associations and boil down to complex
data fitting. Judea Pearl has pointed out that Artificial
General Intelligence must involve interventions involving
the acts of doing and imagining. Any machine assisted
scientific discovery thus must include causal analysis and
interventions. In this context, we propose a causal
learning model of physical principles, which not only
recognizes correlations but also brings out causal
relationships. We use the principles of causal inference and
interventions to study the cause-and-effect relationships
in the context of some well-known physical phenomena. We 
show that this technique can not only figure out associations 
among data, but is also able to correctly ascertain the 
cause-and-effect relations amongst the variables, thereby 
strengthening (or weakening) our confidence in the proposed 
model of the underlying physical process. 
\end{abstract}
\maketitle
\section{Introduction}
Artificial Intelligence (AI), specifically through its
Machine Learning (ML) form, has been successfully applied to
a wide range of fields including agriculture, social media,
gaming, and robotics~\cite{Liakos2018,Angra2017}.  ML plays
a significant role in autonomous driving, natural language
processing, finance, health care, understanding the human
genome, manufacturing, energy harvesting, and much
more~\cite{Angra2017,ml_apps}.

ML has also lent a hand to the scientific community and
has found quite a few applications in scientific research.
In physics specifically, ML has been used to explore many-body
physics~\cite{carleo2017solving,mohseni2021learning}, 
glassy dynamics~\cite{schoenholz2016structural}, 
learning phases of matter~\cite{carrasquilla2017machine,ch2017machine,tibaldi2023unsupervised},
designing new experiments~\cite{krenn2016automated, gao2020computer, babazadeh2017high, erhard2018experimental},
to interpret nature~\cite{friederich2021intuition,shepherd2021interpretable}, 
quantum foundations~\cite{bharti2019teach}, 
quantum state tomography~\cite{torlai2018neural,beach2019qucumber}, 
phase transition~\cite{wang2016discovering,hu2017discovering},  
quantum matter~\cite{carrasquilla2020machine}, 
Monte Carlo simulation~\cite{huang2017accelerated}, 
polymer states~\cite{wei2017identifying},
topological codes~\cite{torlai2017neural}, 
the study of black hole detection~\cite{abbott2016observation},
quantum circuit optimization 
and control ~\cite{fosel2021quantum, fosel2018reinforcement}, 
anti-de Sitter/conformal field theory (AdS/CFT) 
correspondence~\cite{hashimoto2018deep}, 
quantum state preparation~\cite{bukov2018reinforcement1,bukov2018reinforcement2}, 
thermodynamics \cite{torlai2016learning}, 
gravitational lenses~\cite{hezaveh2017fast}, 
characterizing the landscape of string theories~\cite{carifio2017machine}, 
and wave analysis~\cite{biswas2013application}, 
to name a few. An important aim for machine-assisted
scientific discovery, proposed in the seminal work by
Iten \etal where they propose a neural network architecture
modeled after the human physical reasoning 
process~\cite{iten2020discovering}.

\begin{figure}[h]
\hspace*{1cm}\includegraphics[scale=1]{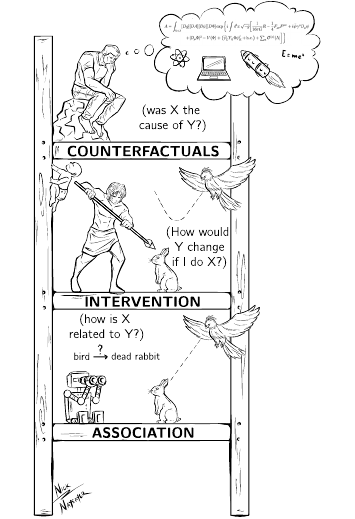}
\caption{The Ladder of Causation depicting the 3 levels
of cognitive ability. Present day ML is at `association',
the lowest level. A machine capable of understanding causal
structures would be placed at `intervention' level
while the more sophisticated AI will also operate at
the level of contrafactuals}
\label{ladder}
\end{figure}

The currently prevalent ML architectures primarily identify
correlations and associations in data and thus the models
only uncover direct connections in the data. Based on the
associations one must learn the causal model and the general
AI systems should be able to uncover the underlying causal
structures.  Therefore, to fully realize the potential of
artificial general intelligence, one needs to incorporate
the essence of cognition within the scope of ML. Judea
Pearl~\cite{pearl2018why} divides this cognitive ability
into three distinct levels as depicted in Fig.~\ref{ladder},
distinguished by the type of query being answered, the
levels are termed as: Association, Intervention, and
Counterfactuals.

The first level(association) of the ladder described in 
Fig.~\ref{ladder} involves predictions based on passive 
observations of the data, \ie data-centric search for 
correlations, associations, regularities and patterns. This 
level answers queries pertaining to observations as to what 
can be found in the data. The second level(intervention),
involves analysis of response to change in variables. 
Rather than just observing the data, one queries the effect
of an induced change, and thus one is looking at a 
cause-and-effect relationship in the variables of the data.
The final level utilizes the causal structure to estimate 
portions of data that don't exist or cannot be observed.
It answers queries related to the hypothetical questions 
that one may imagine - the ``what if'' questions. 
Therefore this level involves Counterfactuals.

One thus sees that most applications of ML in science are
basically at the first level of the ladder. For example, in
the context of the spring-mass vibrating system, ML can find
the relationship between the length of the spring and the
weight attached to it. However ML models cannot answer the
question, ``is the change in spring length caused by the
change in weight or vice-versa''.  Causal Inference takes us
a step above on the ladder of causation and lets us answer
such questions. Once armed with the knowledge of causal
relations, one can begin exploring the counterfactuals
leading to a framework which then becomes a motif for
formulating the laws of nature. Posed a bit differently:
``Had the weight on this spring doubled, its length would have
doubled as well'' (Hooke’s law). - Judea Pearl~\cite{pearl2018why}.

We begin by studying the basics of causal discovery and
causal inference in Section-\ref{causal_inference}. In
Section-\ref{examples} we analyze the causality relations of
some physical phenomenon.  The examples that we consider
include tide height, Ohm's law, light dependent resistance (LDR)
characteristics, and quantum measurement correlations. 
Finally, we close with a discussion on the results 
and possible paths ahead, in Section-\ref{future}.
\section{Causal discovery and inference}
\label{causal_inference}
Causal inference refers to the process of answering
questions based on the underlying causal model of the
cause-and-effect relationship between different variables of
the data. As seen from the ladder of causation
(Fig.~\ref{ladder}), causality relates to response to
interventions. We do a certain action and observe a certain
response.

The limitations of correlations and importance of causal
relations can be easily understood from a simple experiment
of the atmospheric pressure reading of a
barometer~\cite{pearl2018why}. While there is a direct
correlation between the barometer reading and pressure, this
correlation cannot in itself establish the causal
relationship. Is it the barometer reading that causes the
atmospheric pressure to change or is it the atmospheric
pressure that causes the barometer reading to change?  One
requires the knowledge of causal relations to conclude that
it is the pressure that causes the change in reading leading
to the observed correlation and not the other way around.

\begin{figure}[h]
\includegraphics[scale=1]{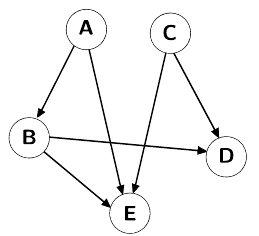}
\caption{A directed acyclic graph with nodes representing
variables and arrows showing the cause-and-effect
relationship between variables.}
\label{dag}
\end{figure}

Statistical algorithms are used to infer the causal
structure from observational data. The model is assumed to
be \textbf{acyclic} where a Directed Acyclic graph(DAG) can
be used to depict the causal relationships as shown in
Fig.~\ref{dag}. The nodes represent the variables and arrows
depict the cause-and-effect relations. The model is
considered to be \textbf{Markovian} where a given node is
conditioned on its immediate parents only.  The model is
assumed to satisfy the conditions of \textbf{Sufficiency}
and \textbf{Faithfulness} which respectively mean that there
exists no external common cause to any pair of nodes in the
graph and all conditional independences (from the underlying
distribution) are completely represented in the graph.  Most
algorithms for causal discovery work with the assumption
that statistical independence implies the absence of causal
relation~\cite{glymour2019review}. Specifically, the Peter
Spirtes and Clark Glymour(PC) algorithm uses the conditional
independence testing criterion to generate a DAG from a
fully connected graph~\cite{kalisch2012causal}, while the
Greedy Equivalence Search(GES) algorithm applies a greedy
search in the graph space to fill an empty graph while
maximizing a fitness measure~\cite{chickering2003optimal}.
Exploiting the asymmetries in models, LiNGAM (Linear
non-Gaussian Acyclic Models) prioritizes the models that
better fit a Linear Non-Gaussian relation among the
variables~\cite{shohei2006linear}. The final goal of causal
discovery process is to arrive at the DAG from the given
data set.

The standard statistics works with correlations which means
working with probability of $Y$ given $X$ denoted by
$P(Y|X)$. Causal inference on the other hand works with
probability of $Y$ given that $X$ is done denoted by
$P(Y|do(X))$ - the do-calculus~\cite{pearl2018why}.  This
`do', though a small change from statistics, is the
representation of an intervention. The difference from
standard ML predictions is that here we are approximating
the effect of treatment $X$ on the outcome $Y,$ based on
data that does not exist in the data-set.

\begin{figure}[h]
\includegraphics[scale=1]{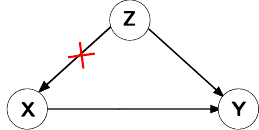}
\caption{Basic aim of causal inference is
to estimate the effect of a treatment $X$ on the outcome $Y$
while controlling for the confounding variables $Z$}
\label{CI_basics}
\end{figure}

The basic idea behind causal inference is to estimate the
effect of treatment $X$ on the outcome $Y$  while
eliminating the dependence on any variable $Z$ that has a
direct influence on both the treatment and the outcome
(confounding variables). This is schematically explained in
Fig.~\ref{CI_basics}. Many methods exist for the estimation
of the effect that is produced by doing $X$. These include
observational studies (conducting and simulating randomized
experiments), simple natural experiments, instrument
variables (specific causal effect estimation criteria) and
refutations. These methods are explained in detail in
Reference~\cite{sharma2018web}.

We use Causal Discovery Toolkit
(CDT)~\cite{kalainathan2019causal} to obtain causal models
directly from the data and DoWhy, ``An end-to-end library
for causal inference''~\cite{sharma2020dowhy}, to carry out
the causal analysis.

\noindent The basic analysis involves the following steps:
\par
\noindent{\bf 1. Creating a causal model:}
We create an initial model of the phenomena that we are
studying as a Directed Acyclic Graph. The DAG is input into
the DoWhy library as a dot graph (a textual representation
of the graph using DOT Language)~\cite{gansner2006dot}.
This initial model is either extracted from the data using
CDT or from domain knowledge.

\par
\noindent{\bf 2. Causal effect identification:} 
Based on causal model, we identify the causal effects to be
estimated using a suitable criterion 
among the following:
\begin{itemize}
\item[\bf a:] {\bf Back-door:} Controlling for the set of
variables that block all the back-door paths between the
treatment and the outcome. A Back-door path is any path
connecting treatment to outcome via an arrow inward
on the treatment. In Fig.~\ref{causal_model},
$X\leftarrow Z\rightarrow Y$ is a back-door path from
treatment $X$ to outcome $Y$. Adjusting for the variable $Z$
will be the back-door criterion: $$ P(Y|do(X)) = \sum_z
P(Y|X,z)P(z) $$

\begin{figure}[h]
\includegraphics[scale=1]{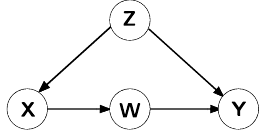}
\caption{A sample causal model as a DAG. $X$
is the treatment, $Y$ the outcome, $Z$ a confounder, and
$W$ a mediator. $X\leftarrow Z\rightarrow Y$ constitutes
the backward path while $X\rightarrow W\rightarrow Y$ is
the forward path from treatment to outcome}
\label{causal_model}
\end{figure}

\item[\bf b:] {\bf Front-door:} Controlling of variables
in the forward path from the treatment to outcome. In
Fig.~\ref{causal_model}, $X\rightarrow W\rightarrow Y$ is
the front-door path from treatment to outcome. Adjusting 
for variables $X$ and $W$ will be the front-door criterion:
$$ P(Y|do(X)) = \sum_w P(w|X) \sum_{x}P(Y|x,w)P(x) $$

\item[\bf c:] {\bf Instrumental
variables~\cite{sharma2018web}:} A special case of the
front-door criteria, this method helps in identifying the
direct causal estimate from $X$ to $Y$ when the back-door
criterion fails (e.g. - obtaining data on $Z$ is not
possible, and hence $Z$ cannot be controlled for).  This
method can only be  applied if there exists a variable which
is independent of confounders of treatment and outcome, has
a direct relation with the treatment, and has no direct
effect on the outcome as depicted in Fig.~\ref{iv_model}.

\begin{figure}[h]
\includegraphics[scale=1]{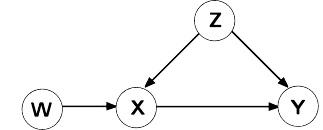}
\caption{Causal model for Instrumental Variable.
Since $W$ is independent of confounder $Z$, and has direct effect on
treatment $X$, and has no direct effect on the outcome $Y$,
it can be used to estimate the effect of $X$ on $Y$ (given that
controlling for $Z$ is not possible)}
\label{iv_model}
\end{figure}

\item[\bf d:] {\bf Mediation:}
This method is applied when the treatment has multiple
causal pathways to the outcome as shown in Fig.~\ref{mediation_model}.
It enables us to separate the total effect on $Y$ into direct 
($X\rightarrow Y$) and indirect ($X\rightarrow W\rightarrow Y$)
causal estimates.

\begin{figure}[h]
\includegraphics[scale=1]{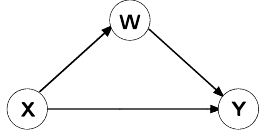}
\caption{Mediation causal model. The treatment
$X$ has two causal pathways to the outcome $Y$, direct
($X\rightarrow Y$) and indirect
($X\rightarrow W\rightarrow Y$) via the mediator $W$}
\label{mediation_model}
\end{figure}
\end{itemize}

\par
\noindent{\bf 3.~Estimate the target estimand:}
Many statistical methods exist for estimating the
indentified casal effect. Depending on the identification
criteria one can use linear regression, distance matching,
propensoty score stratification \cite{rosenbaum1983central}
for backdoor; wald estimator \cite{wald1943test}, regression
discontinuity \cite{thistlethwaite1960rdd} for instrument
variable; two-stage linear regression for frontdoor and so
on. The estimate is obtained in units of Average Treatment
Effect (ATE), Average treatment effect for the treated (ATT)
or Average treatment effect for the controls (ATC).

\par
\noindent{\bf 4.~Refute the obtained estimate using multiple
robustness checks:} 
Causal models are not absolute, as  they cannot be proven to
be correct or incorrect. One can however, increase faith in a
model by checking the validity of the assumptions behind the
model against various robustness checks which include:
\begin{itemize}
\item[\bf a:]
{\bf Random Common Cause:}
Check the variation of estimate over addition of an independent
random common cause. Lesser the variation, higher our faith
in the model.
\item[\bf b:]
{\bf Placebo Treatment Refuter:} Rerunning the analysis with
an independent random variable as the treatment variable. If
the initial treatment is in fact the cause, the new estimate
should go to zero.
\item[\bf c:]
{\bf Data Subset Refuter:} How much is the variation in
estimate when only a subset of the data is used? The
variation is small for a strong causal relation.
\end{itemize}
\section{\label{examples}{EXAMPLES}}
This section describes our main work where we have chosen
four different examples to build causal models. For each
case we consider different possible causal models and evaluate
their relative efficacy by employing the methods described 
above. The examples are chosen from diverse fields. The 
first example of tides and cause of their varying height 
over the year is about a natural phenomenon where data is taken
from documented sources. The second example is about a physics
model involving Ohm's law and direct and indirect dependence 
of current on various possible parameters. The third example is
about an actual experiment where we collect data for a light
dependent resistance(LDR) and consider various possible causal
models for it, which we evaluate and compare using data and 
domain knowledge. In the last example we consider quantum
correlations in the two-party two-value setting, and ask the 
question as to what is the most plausible cause of these 
non-trivial quantum correlations.
\subsection{Height of Tides}
\label{tide_section}
It is a well known fact that tides, the rise and fall of sea
levels, are a cumulative effect of Sun and Moon's
gravitational force on Earth, among other minor factors. We
ask a ML model, which of the two - {\bf Sun} or {\bf Moon} -
plays a bigger role in determining the maximum height of the
tide on a given day. To that end, we prepare a data-set with
daily Earth-Sun distance in astronomical units(AU),
Earth-Moon distance (in AU), and the maximum height of tide
at four different locations- Honolulu (Hawaii), Mumbai
(India), Liverpool (England), Halifax (Canada)

\begin{figure}[h]
\includegraphics[scale=1]{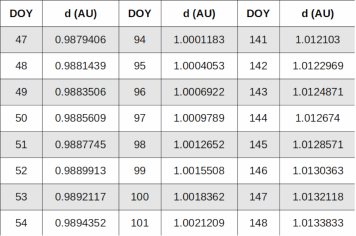}
\caption{Sample Earth-Sun distance data. The data includes
day of the year ({\bf DOY}) and the Earth-Sun distance {\bf
d}(in AU)}
\label{sample_ES}
\end{figure}

We collected the year round data of earth-moon distance,
earth-sun distance, and tide height from documented sources.
The {\bf earth-sun} distance for a given day of the year is
obtained from the csv file available on the
\href{https://www.usgs.gov/media/files/earth-sun-distance-astronomical-units-days-year}{USGS
webpage}. A sample of the dataset is shown in
Fig.~\ref{sample_ES}

\begin{figure}[h]
\includegraphics[scale=1]{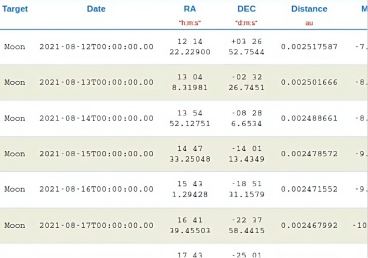}
\caption{Sample Earth-Moon distance data, 2019. The website
generates the ephemeris data for the Moon.}
\label{sample_EM}
\end{figure}

\begin{figure}[h]
\includegraphics[scale=1]{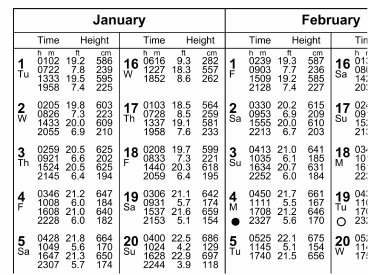}
\caption{Sample Tidal data for Liverpool, England, 2019.
Each box of a given date records the time (in hours and
minutes) height of high and low tides (in feet and cm)}
\label{sample_tides}
\end{figure}

The {\bf earth-moon} distance is extracted from
\href{http://vo.imcce.fr/webservices/miriade/?forms}{IMCCE
VIRTUAL OBSERVATORY}. Fig.~\ref{sample_EM} shows the sample
of the table generated at the webpage.
The data of {\bf tide height} is available as PDF file on the
\href{https://tidesandcurrents.noaa.gov/historic\_tide\_tables.html}{NOAA
website}. Fig.~\ref{sample_tides} shows a sample of the PDF.
At any given day, the tide heights were recorded 3-4 times.
We used the maximum value of height (in ft) for a given day.

\begin{figure}[h]
\begin{subfigure}[b]{0.18\textwidth}
\includegraphics[scale=1]{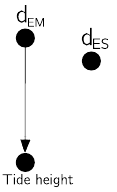}
\subcaption{From LiNGAM algorithm in CDT}
\end{subfigure}
\hspace{10mm}
\begin{subfigure}[b]{0.18\textwidth}
\includegraphics[scale=1]{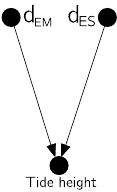}
\subcaption{From Domain Knowledge}
\end{subfigure}
\caption{Causal diagram for tide height. $d_{ES}$ and
$d_{EM}$ represent the earth-sun and earth-moon distance
respectively}
\label{tides_model}
\end{figure}

We prepared the models as described in
Fig.~\ref{tides_model} and computed two causal estimates
with tide-height as the outcome. We used the Earth-Moon
distance as target for the first estimate and Earth-Sun
distance for the second.

The causal diagram predicted from data only gives us the
$d_{EM}\rightarrow h$ causal relation, which is in fact the
most significant one. The estimates (in ATE) from the
predicted and ground-truth models differ marginally:
-2964.45 and -2913.16 respectively (for Halifax). The
causal-estimates for Earth-Moon and Earth-Sun distance (in
ATE) for the ground-truth model are listed in 
table~\ref{table_tides}.
\begin{table}[h]
\begin{tabular}{|c|c|c|c|c|}
\hline
\multirow{2}*{Causal Relation} &
\multicolumn{4}{|c|}{Estimate (ATE)}\\
\cline{2-5}
& Halifax & Liverpool & Honolulu & Mumbai\\
\hline
$d_{EM}$ $\rightarrow$ h & -2913.16 & -10045.83 & -1205.91 & -7232.59\\
$d_{ES}$ $\rightarrow$ h & -2.20 & -8.62 & -3.34 & -22.15\\
\hline
\end{tabular}
\captionof{table}{Causal estimates for $d_{EM}\rightarrow h$
and $d_{ES}\rightarrow h$ at the four locations. The ATE
values are shown for Earth-Moon and Earth-Sun for all four
locations.}
\label{table_tides}
\end{table}
It is clearly visible from the estimates and from the causal
diagram obtained from data, that the Earth-Moon distance is the
primary cause for the tide height. 
\subsection{Ohm's law}
\label{ohm_section}
\begin{figure}
\hspace*{-0.5cm}
\begin{subfigure}[b]{0.15\textwidth}
\includegraphics[scale=1]{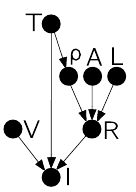}
\subcaption{including faux-common-cause (T)}
\label{ohm_model_all}
\end{subfigure}
\begin{subfigure}[b]{0.15\textwidth}
\includegraphics[scale=1]{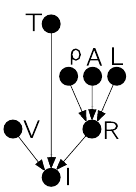}
\subcaption{false relation model - rejected while refuting}
\label{ohm_model_false}
\end{subfigure}
\begin{subfigure}[b]{0.15\textwidth}
\includegraphics[scale=1]{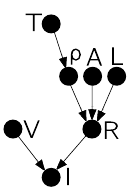}
\subcaption{true relation model - no refutations}
\label{ohm_model_true}
\end{subfigure}
\caption{Causal diagram for Ohm's law. For a wire with
potential $V$ across its length $L$, resistivity $\rho$,
cross-section area $A$ at temperature $T$ }
\label{ohm_model}
\end{figure}

In this example, we look for the driving forces (cause) of 
the current $I$ in a wire of length $L$, cross-sectional area
$A$, resistivity $\rho$, at a temperature $T$, with a potential
$V$ applied across its ends.  Using causal analysis one can
test the validity of a given cause and effect relation. To 
that end, we consider and check the validity of a model with
a direct $T\rightarrow I$ arm added in addition to the known
dependence of $I$ on $T$ via $R$.  Different causal-models
that we evaluate are depicted in Fig.~\ref{ohm_model}.

\begin{figure}[h]
\includegraphics[scale=1]{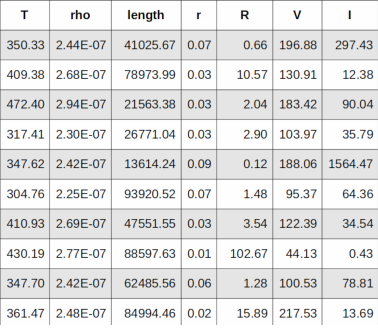}
\caption{Sample of the data-set used in the analysis.
Current $I$ resulting from Potential $V$ applied across a
wire of length $L$, resistance $R$ (resistivity $\rho$,
cross-section area $A$) at temperature $T$}
\label{sample_ohm}
\end{figure}

Using the known relations (Eq.~(\ref{ohm_law_eqn})) between
current and voltage, and the temperature dependence of
resistance, we generate the required data. We use platinum
as the material for our constants ($\alpha,\rho_0$).
Fig.~\ref{sample_ohm} shows a sample input data.
\begin{gather}
\nonumber V = I R\\
\nonumber R = \frac{\rho_tL}{A}\\
\rho_t = \rho_0 (1+\alpha\Delta T)
\label{ohm_law_eqn}
\end{gather}

The candidate causal models depicted in Fig.~\ref{ohm_model}
are evaluated and estimates in terms of ATE values are
computed which are tabulated in Table~\ref{table_ohm}.
We see that the major driving force is potential $V$, with
resistance showing an inverse relation as expected.

We observe that the effect of $T$ on $I$ is not only
non-zero, but equivalent to that of $R$. The fact that this
effect follows not from the direct $T\rightarrow I$ path,
but the $T\rightarrow \rho \rightarrow R\rightarrow I$ path
is confirmed by estimating the same effect of $T$ on $I$
using a causal model which does not have the $T\rightarrow
I$ path (Fig.~\ref{ohm_model_true}). We get the same ATE
value of 0.218.

One can also check the effect by removing the other branch:
$T\rightarrow\rho$ (Fig.~\ref{ohm_model_false}). This results
in an estimate(ATE) of 1.35, but during the placebo treatment
refutation, the new estimated effect in terms of ATE values,
which should be 0, comes out to be -10.54 and thus shows 
that this model is less trustworthy. 
\begin{table}[h]
\begin{tabular}{|c|c|c|c|}
\hline
\multirow{2}*{Causal Relation } &
\multicolumn{3}{|c|}{Estimate (ATE)}\\
\cline{2-4}
& Model A & Model B & Model C\\
\hline
V $\rightarrow$ I & 1.735 & 1.735 & 1.735\\
R $\rightarrow$ I & -0.205 & -0.225 & -0.205\\
T $\rightarrow$ I & 0.218 & 1.35 & 0.218\\
\hline
\end{tabular}
\caption{\label{table_ohm}Causal estimates for different
causal relations in the three models of Ohm's Law}
\end{table}
\subsection{Power and LDR Resistance}
\label{ldr_exp}
\begin{figure}[h]
\includegraphics[scale=1]{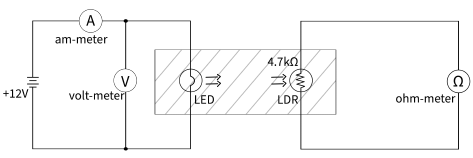}
\caption{Circuit Diagram for the LDR Experiment.
The LED and LDR are placed in a closed box at a fixed
distance from each other. The LED is supplied with a variable
voltage. The voltmeter measures the voltage across the LED,
the ammeter measures the current through the LED, and the
ohmmeter measures the resistance of LDR. The experiment is
repeated with flux meter in place of LDR to obtain power
readings.
\label{circuit_diagram}
}
\end{figure}

Next we perform causal analysis of real data obtained
from an experiment. A light emitting diode(LED) light
source, runs using a battery at voltage $V$ and draws
current $I$. The light emitted by the LED shines on and
light dependent resistance(LDR) and this  provides power $P$
to LDR thereby changing its resistance $R$.  The circuit is
described in Fig.~\ref{circuit_diagram}. 

\begin{figure}[h]
\begin{subfigure}[b]{0.18\textwidth}
\includegraphics[scale=1]{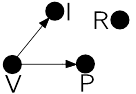}
\caption{From LiNGAM algorithm in CDT}
\label{ldr_model_a}
\end{subfigure}
\begin{subfigure}[b]{0.24\textwidth}
\includegraphics[scale=1]{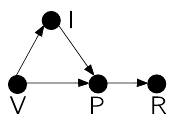}
\caption{From Domain Knowledge}
\label{ldr_model_b}
\end{subfigure}
\caption{Causal diagrams for the LDR experiment.}
\label{ldr_model}
\end{figure}

\begin{table}[h]
\begin{tabular}{|c|c|c|c|}
\hline
Voltage & Current & Power & Resistance \\
(V) & (mA) & (lux) & (k$\Omega$) \\
\hline
2.67 & 100.3 & 5 & 37.000 \\
2.90 & 104.7 & 9 & 25.400 \\
3.17 & 109.9 & 15 & 17.800 \\
3.68 & 119.2 & 36 & 6.800 \\
3.84 & 122.2 & 47 & 5.790 \\
\vdots & \vdots & \vdots & \vdots \\
7.06 & 172.0 & 847 & 0.591 \\
7.22 & 174.1 & 923 & 0.558 \\
7.70 & 179.2 & 1158 & 0.459 \\
7.86 & 181.5 & 1266 & 0.435 \\
8.00 & 183.6 & 1386 & 0.413 \\
\hline
\end{tabular}
\captionof{table}{Sample data from the experiment.
At each voltage setting, the current through LED is measured in mA
and the LDR's resistance is measured in k$\Omega$. The experiment
is repeated with LDR replaced by a flux meter to measure power.}
\label{table_ldr_data}
\end{table}

The model obtained from data (Fig.~\ref{ldr_model_a})
suggests potential
$V$ as the cause for both power $P$ and current $I$, and finds no
direct causal relation between power $P$ and the LDR resistance $R$.
The model as expected provide only the most significant cause-effect
relations (Table \ref{table_ldr_estimates}). We know, as depicted in the
domain knowledge model (Fig.~\ref{ldr_model_b}), that current $I$ acts as
a mediator for $V$'s effect on $P$. This becomes clear when we compare
refutations of models with and without $I\rightarrow P$ arm.
Refutations suggest that we put more faith in the model with $I\rightarrow P$ arm
(p-value 0.912) than the ones without this arm (p-value 0.882)
(Table \ref{table_ldr_confidence}). The analysis also
suggest that we put more faith
in the model which includes the $P\rightarrow R$ arm over the one
that does not contain this arm.
\begin{table}[h]
\begin{tabular}{|l|c|c|c|}
\hline
\multirow{2}*{Model} & \multicolumn{3}{|c|}{Estimates (ATE)}\\
\cline{2-4}
& $V\rightarrow P$ & $V\rightarrow I$ & $P\rightarrow R$\\
\hline
Data & 251.533 & 15.42 & -\\
\hline
Domain & \multirow{2}*{251.533} & \multirow{2}*{15.41} 
& \multirow{2}*{-0.008}\\
Knowledge & & &\\
\hline
\end{tabular}
\caption{Causal estimates of three causal relations
for data and the domain knowledge models in the LDR experiment}
\label{table_ldr_estimates}
\end{table}

\begin{table}[h]
\begin{tabular}{|c|c|c|}
\hline
 & $P\rightarrow R$ present & $P\rightarrow R$ absent\\
\hline
$I\rightarrow P$ present & 0.928 & 0.896\\
\hline
$I\rightarrow P$ absent  & 0.897 & 0.867\\
\hline
\end{tabular}
\caption{Confidence levels of $V\rightarrow P$
causal estimate for different causal models in the LDR experiment}
\label{table_ldr_confidence}
\end{table}
\subsection{Measurement correlation and quantum entanglement}
\label{entang}
The last example we choose is from the domain of quantum
mechanics.  Quantum states of composite systems can show
peculiar kinds of correlations.  We analyze these
correlations from the point of view of constructing a causal
model. 

\begin{figure}[h]
\includegraphics[scale=1]{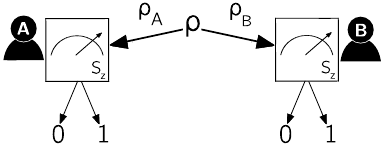}
\caption{Alice and Bob with a shared  quantum state $\rho$ of two spin
half particles. Each measures the spin of their particle along
the $z$-axis and gets one of two possible outcomes (0 or 1).}
\label{alice_bob}
\end{figure}

A quantum spin half particle is a two level quantum system with
its state space consisting of normalized densities over a 
two dimensional complex linear vector space.
\cite{cho2022two-level} The measurables 
for each particle are spin components in any direction and the
spin component takes two values  `up'(1) or `down'(0) when
measured and are the eigen values of the corresponding
Hermitian operator. For example if we are measuring the $z$
component of spin the corresponding observable is the Pauli
matrix $\sigma_z$.  The scenario that we consider consists of
two spin half particles which are in a joint quantum state
$\rho$. Alice and Bob are two observers with the capability of
measuring spin components and the first particle is
accessible to Alice while the second is accessible to Bob. The
scenario is schematically  depicted in Fig.~\ref{alice_bob}.

Consider the case where both Alice and Bob 
measure the spin of their respective particle along the 
$z$-axis which corresponds to measuring the operator
$\sigma_z$ in the appropriate state space. For each the
possible outcomes thus are $0$ or $1$. Therefore,
the joint measurement outcome for the composite system
will be in the set \{00,01,10,11\}.  One can compare this
situation to the one of tossing two coins in the classical
domain where the outcome set is the same and if the coins are
unbiased, the probability of each outcome will be equal.

Quantum states have a property called quantum
entanglement~\cite{horodecki2009entanglement}
which is considered to be responsible for unusual
correlation properties of composite quantum systems. The
entanglement can be  mathematically computed from the given
density operator and can be quantified via a measure called
log-negativity~\cite{plbnio2007entanglement_measures}. For 
certain maximally entangled states the outcomes can be such 
that they always either fall in  the set \{01,10\} or the set
\{00,11\}, {\it i.e.} the outcomes are always (anti)correlated.
This scenario is schematically described in Fig.~\ref{alice_bob}.
The data set the we analyze is generated by simulating the
measurement setup between Alice and Bob.  A state of the
composite  quantum system $\rho$ is generated randomly. On
this randomly generated state, both Alice and Bob, perform a
$\sigma_z$ measurement. They repeat these measurements on
the state  100 times and these 100 measured values are used
to compute the correlation. The entanglement is computed from the
state density matrix mathematically by computing the
log-negativity. This process is repeated for another
randomly generated composite state of the two spins.  Twenty
such random states are chosen and thus a data-set with 2000
rows is generated with 100 rows corresponding to a given
random state $\rho$.  The data set is schematically
described in Table~\ref{q-data}. As can be seen for each
$\rho$ we have 100 rows which are used to calculate the
correlations and have mathematically computed
log-negativities as documented in the second column.
\begin{table}[h]
\begin{tabular}{|c|c|c|c|c|c|}
\hline
State&Entanglement& $M_A$ & $M_B$ &Correlation& 
\multirow{5}{*}{
\begin{tabular}[c]{l}\\
\shortstack{instance 1\\ (100 rows)}
\end{tabular}
 } \\
\hline
$\rho_1$ &  $E_1$  & +1   & -1   & $C_1$ & \\
..       &  ...    & ..   & ..   & ...   & \\
$\rho_1$ &  $E_1$  & -1   & -1   & $C_1$ & \\
&&&&&\multirow{5}{*}{
\begin{tabular}[c]{l}\\
\shortstack{instance 2\\ (100 rows)}
\end{tabular}
 } \\
\hline
$\rho_2$ &  $E_2$  & +1   & +1   & $C_2$ & \\
..       &  ...    & ..   & ..   & ...   & \\
$\rho_2$ &  $E_2$  & -1   & +1   & $C_2$ & \\
\hline
..       &  ...    & ..   & ..   & ...   & \\
..       &  ...    & ..   & ..   & ...   & \\
&&&&&\multirow{5}{*}{
\begin{tabular}[c]{l}\\
\shortstack{instance $n$\\ (100 rows)}
\end{tabular}
 } \\
\hline
$\rho_n$ &  $E_n$  & -1   & +1   & $C_n$ & \\
..       &  ...    & ..   & ..   & ...   & \\
$\rho_n$ &  $E_n$  & +1   & +1   & $C_n$ & \\
\hline
\end{tabular}

\caption{Structural setup of the data generated
in the simulation of Alice and Bob's $\sigma_z$
measurements on two  entangled spin half particles.
A single instance is 100 samples of measurements performed by
Alice ($M_A$) and Bob ($M_B$) on the shared state $\rho$. The
correlation value $C$ is evaluated from these 100 samples.
\label{q-data}
}
\end{table}

Where is the cause of the correlation between the measured
values of Alice and Bob?
The initial causal discovery attempts failed to reveal any
relations between the variables. Upon further investigation
into the data, we find that the present scenario is a
special case where the variables of interest, though
causally linked (as we know from Domain knowledge), have
zero correlation between them. Entanglement ranging from 0
to 1, while Correlation ranging from -1 to +1 creates a case
where the average correlation between these two variables is
(very close to) zero.

Tackling such cases involves looking at the causal relation
among functions of the involved variables. We take the
absolute value of correlation as the second variable of
interest and continue with the analysis.
\begin{figure}[h]
\begin{subfigure}[b]{0.18\textwidth}
\includegraphics[scale=1]{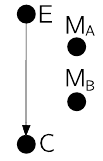}
\caption{From LiNGAM algorithm in CDT}
\end{subfigure}
\begin{subfigure}[b]{0.24\textwidth}
\includegraphics[scale=1]{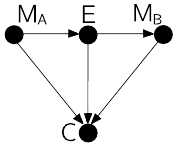}
\caption{From Domain Knowledge (with additional
assumption)}
\end{subfigure}
\caption{Causal diagram for correlation between Alice and
Bob's measurement outcomes ($M_A$ and $M_B$ respectively).}
\label{entanglement_model}
\end{figure}
Similar to the tide-height example, the predicted causal
diagram only shows the most significant cause-effect
relation. With $M_A$ as the treatment and correlation ($C$)
as the outcome (while accounting for entanglement($E$) as a
common cause), we get a causal estimate of -0.0002 ATE.
Similarly, we get -0.0024 ATE for $M_B$
(Table~\ref{table_correlation}).

The estimate for entanglement $E$ as the treatment is 0.3733
ATE.  This shows that the machine puts more faith in the
model that has Entanglement ($E$), as the underlying
variable as the cause of the correlation ($C$) between $M_A$
and $M_B$ over the model that assumes either $M_A$ or $M_B$
as the cause for $C$.  Therefore, the ML analysis confirms
this fact that we know from the domain knowledge.

\begin{table}[h]
\begin{tabular}{|c|c|c|}
\hline
\multirow{2}*{Causal Relation} & Estimate & Confidence \\
& (ATE) & (p-vale) \\
\hline
$M_A\rightarrow C$ & -0.0002 & 0.82\\
$M_B\rightarrow C$ & -0.0024 & 0.72\\
$E\rightarrow C$ & 0.3733 & 0.94\\
\hline
\end{tabular}
\caption{Causal estimates for measurement correlation and
entanglement. The p-values are averaged over three different
refutation methods.}
\label{table_correlation}
\end{table}

This statement is further strengthened by the results of
refuting the causal model in both the above scenarios. For
example, the placebo treatment refutation gives a confidence
of 94\% (p-value: 0.94) in the former model and a p-value of
$\sim$0.8 for the latter.
\section{Discussion and Future Work} 
\label{future}
While standard AI and ML based techniques have an
outstanding performance in association level tasks, they are
unable provide answers to basic queries of cause and effect.
One requires the use of causal diagrams and causal inference
to equip the machine with said capability. Association level
inference is possible even for a machine that understands
cause-and-effect relation.  Section~\ref{tide_section} shows
that using causal analysis framework one can infer that
while both Sun and Moon's gravitational pull affects the
tides on Earth, the Earth-Moon distance is the major cause
for the height of tides.  The advantage of causal models
over simple associative models is seen in
Section~\ref{ohm_section}. Not only does the machine
estimates potential to be the primary cause for current, it
refutes the incorrect assumption of temperature {\it
directly} affecting the current. The LDR experiment analysis
shows that, while the conclusions are not 100\% accurate,
using causal discovery to infer causal relations from 
experimental data can hint at where the focus in the 
experiment should be. In the problem related to
cause of correlations between quantum measurements, we
observe that the machine is able to figure out the
underlying cause of the correlation between the
measurement outcomes of Alice and Bob being the quantum
entanglement.

Causal Theory is still in its initial stages of development
and therefore is in no way foolproof. There exist quite a
few different algorithms for causal discovery and there is
no guarantee of the outcomes of one agreeing with the
outcomes of the other. The approach is data-centric and does
not always yield relations that make sense. Nonetheless,
having an initial estimate of a causal model helps speed up
the process. One can always fine-tuned the estimates and
relations using domain knowledge. Adding the layer of causal
analysis can deepen the understanding of the phenomena and
processes involved. The authors in
\cite{udrescu2020aifeynman} present a physics-inspired
symbolic regression ML algorithm for discovering
expressions/equations from data alone. One can explore the
advantage of incorporating causal inference to such ML
applications.

\begin{acknowledgments}
J.S. would like to thank Amitoj Kaur Chandi 
(\href{https://www.instagram.com/nick_naysayer/}{@nick$\_$naysayer})
for Fig.~\ref{ladder} and Dr. Paramdeep Singh for help with
the experimental setup for Section \ref{ldr_exp}. 
J.S. acknowledges IISER Mohali for financial support.
\end{acknowledgments}

%
\clearpage
\onecolumngrid

\appendix
\section{Causal Analysis - DoWhy}
DoWhy is ``An end-to-end library for causal inference''
developed by Microsoft.  It
abstracts the entire process of causal analysis into a 4
step process: model, identify, estimate, and refute.

\noindent Where the general trend of causal inference is
estimation of a model parameter like the coefficient of
linear regression, DoWhy provides a do-sampler that
estimates the distribution of $P(Y|do(X=x))$. This enables
us to compute statistics other than test and control
difference in average outcomes.

\noindent Considering the example of tide heights:

\begin{enumerate}
\item Model

One can provide a model as a digraph from the knowledge
related to the subject:
\begin{lstlisting}
causal_graph = ```digraph {
EMd[label=``Earth-Moon distance''];
ESd[label=``Earth-Sun distance''];
h[label=``height of the tide'']
ESd->EMd->h;
ESd->h;
}'''
model= dowhy.CausalModel(data = dataset_halifax,
                        graph=causal_graph.replace(``\n'', `` ''),
                        treatment=`EMd',
                        outcome=`h',
                        common_causes=`ESd')
model.view_model()
display(Image(filename=``causal_model.png''))
\end{lstlisting}
Or one can obtain the model from the data itself using
causal discovery algorithms (CDT package):
\begin{lstlisting}
from cdt.causality.graph import LiNGAM
labels = list(dataset_halifax.columns)
predicted_graph = LiNGAM().predict(dataset_halifax)
adj_matrix = np.asarray(nx.to_numpy_matrix(predicted_graph))
idx = np.abs(adj_matrix) > 0.01
dirs = np.where(idx)
graph = graphviz.Digraph(engine='dot')
for name in labels:
  graph.node(name)
for edge_to, edge_from, value in zip(dirs[0], dirs[1], adj_matrix[idx]):
  graph.edge(labels[edge_from], labels[edge_to], label=str(value))
display(graph)
\end{lstlisting}
\textbf{Result:}
\begin{figure}[h]
\begin{subfigure}[b]{0.40\textwidth}
\includegraphics[scale=0.5]{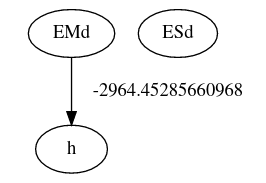}
\caption{From LiNGAM algorithm in CDT}
\end{subfigure}
\begin{subfigure}[b]{0.40\textwidth}
\includegraphics[scale=0.5]{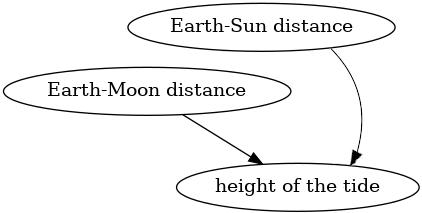}
\caption{From Domain Knowledge}
\end{subfigure}
\end{figure}

\item Identify
\begin{lstlisting}
import statsmodels
identified_estimand = model.identify_effect(
                    proceed_when_unidentifiable=True)
print(identified_estimand)
\end{lstlisting}
\textbf{Result:}
\begin{lstlisting}
Estimand type: nonparametric-ate

### Estimand : 1
Estimand name: backdoor
Estimand expression:
d                   
-------(Expectation(h))
d[ESd]                
Estimand assumption 1, Unconfoundedness: If U->{ESd} and U->h
then P(h|ESd,,U) = P(h|ESd,)

### Estimand : 2
Estimand name: iv
No such variable found!

### Estimand : 3
Estimand name: frontdoor
No such variable found!
\end{lstlisting}

\item Estimate
\begin{lstlisting}
estimate = model.estimate_effect(identified_estimand,
                                method_name=``backdoor.linear_regression'',
                                control_value=0,
                                treatment_value=1,
                                confidence_intervals=True,
                                test_significance=True)
print(``Estimate:'',estimate.value)
\end{lstlisting}
\textbf{Result:}
\begin{lstlisting}
Estimate: -2913.157882942987
\end{lstlisting}

\item Refute
\begin{lstlisting}
# Placebo Treatment Refuter:- Randomly assigns any covariate as
# a treatment and re-runs the analysis. If our assumptions were 
# correct then this newly found out estimate should go to 0.
refute = model.refute_estimate(identified_estimand, estimate,
        method_name=``placebo_treatment_refuter'')
print(refute)
\end{lstlisting}
\textbf{Result:}\\
\begin{lstlisting}
Refute: Use a Placebo Treatment
Estimated effect:-2913.157882942987
New effect:0.0
p value:1.0
\end{lstlisting}

\end{enumerate}

\end{document}